\documentclass[sn-mathphys,Numbered]{sn-jnl}% Math and Physical Sciences Reference Style
\usepackage{graphicx}%
\usepackage{multirow}%
\usepackage{amsmath,amssymb,amsfonts}%
\usepackage{amsthm}%
\usepackage{mathrsfs}%
\usepackage[title]{appendix}%
\usepackage{xcolor}%
\usepackage{textcomp}%
\usepackage{manyfoot}%
\usepackage{booktabs}%
\usepackage{algorithm}%
\usepackage{algorithmicx}%
\usepackage{algpseudocode}%
\usepackage{listings}%

\theoremstyle{thmstyleone}%
\theoremstyle{thmstyletwo}%

\theoremstyle{thmstylethree}%

\begin{document}


\title[Article Title]{No-reference Quality Assessment of Contrast-distorted Images using Contrast-enhanced Pseudo Reference}


\author*[1]{\fnm{Mohammad-Ali} \sur{Mahmoudpour}}\email{mohamadali.mahmoodpour@alumni.sbu.ac.ir}

\author[2]{\fnm{Saeed} \sur{Mahmoudpour}}\email{Saeed.Mahmoudpour@vub.be}



\affil*[1]{\orgdiv{Dept. of Electrical Engineering}, \orgname{Shahid Beheshti University}, \orgaddress{\postcode{1983963113}, \state{Tehran}, \country{Iran}}}

\affil[2]{\orgdiv{Dept. of Electronics and Informatics}, \orgname{Vrije Universiteit Brussel}, \orgaddress{\postcode{1050}, \state{Brussel}, \country{Belgium}}}

\newcommand{\multrow}[1]{\begin{tabular}{@{}c@{}} #1 \end{tabular}}

\abstract{Contrast change is an important factor that affects the quality of images. During image capturing, unfavorable lighting conditions can cause contrast change and visual quality loss. While various methods have been proposed to assess the quality of images under different distortions such as blur and noise, contrast distortion has been largely overlooked as its visual impact and properties are different from other conventional types of distortions. In this paper, we propose a no-reference image quality assessment (NR-IQA) metric for contrast-distorted images. Using a set of contrast enhancement algorithms, we aim to generate pseudo-reference images that are visually close to the actual reference image, such that the NR problem is transformed to a Full-reference (FR) assessment with higher accuracy. To this end, a large dataset of contrast-enhanced images is produced to train a classification network that can select the most suitable contrast enhancement algorithm -- based on image content and distortion -- for pseudo-reference image generation. Finally, the evaluation is performed in the FR manner to assess the quality difference between the contrast-enhanced (pseudo-reference) and degraded images. Performance evaluation of the proposed method on three databases containing contrast distortions (CCID2014, TID2013, and CSIQ), indicates the promising performance of the proposed method.}


\maketitle

\section{Introduction}\label{sec1}

With the increasing use of mobile phones and social media networks, a huge number of images are now available to people. These images may suffer from various degradations during various stages, such as acquisition, compression, and transmission. Internet service providers seek methods to automatically assess the quality of images to optimize the performance of their network components while improving the quality of experience (QoE) to end-users. To this end, image quality assessment (IQA) methods are necessary to assess the quality of images with high consistency to human quality opinions.

IQA methods can be divided into two main categories subjective and objective assessments. In subjective methods, humans are asked to rate the quality of images. While this approach is highly reliable, it is costly and time-consuming, making it unsuitable for real-time applications. Objective IQA algorithms aim to automatically evaluate image quality by emulating how humans perceive images. 

There are three main types of objective IQA methods: Full-reference IQA (FR-IQA), reduced-reference IQA (RR-IQA), and no-reference IQA (NR-IQA). FR-IQA methods assess image
quality by comparing a high-quality reference image with a degraded image and RR-IQA methods require only partial information from a reference image to evaluate quality. NR-IQA methods, on the other hand, do not require a reference image and assess quality solely based on the degraded image. Designing NR-IQA methods is challenging due to the lack of a reference image while it is highly desirable in many applications where a reference image is not available.\\

The most widely-used FR-IQA methods include Peak Signal-to-Noise Ratio (PSNR)\cite{psnr} and Structural Similarity Index (SSIM)\cite{ssim}. While PSNR computes pixel-wise error between a reference image and a degraded image, SSIM, assesses quality by comparing the structural similarity, taking into account the sensitivity of the human visual system (HVS) to structural features. The Multi-Scale SSIM (MS-SSIM)\cite{ms-ssim} method improves upon SSIM by computing SSIM at multiple scales, simulating the multiscale analysis of images by the human visual system. It combines these measurements through weighted summation. 
Visual Information Fidelity (VIF)\cite{vif} method models the reference image as the output of a natural stochastic source that passes through the HVS channel to characterize natural scene statistics(NSS). It then computes similar statistics for the degraded image in the presence of a degradation channel, and the ratio of the obtained information between the reference and distorted image is used as the evaluation score. 
Gradient Magnitude Similarity Deviation (GMSD)\cite{gmsd} method evaluates quality by calculating the pixel-wise similarity of gradient magnitudes between the reference and the degraded image.~DISTS \cite{DISTS} method is based on a mapping function derived from a deep neural network, which combines SSIM-like metrics and texture similarity between feature maps of two images.

The Reduced Reference Entropic Differencing (RRED)\cite{rred} method calculates differences in wavelet coefficient entropies between the reference and the degraded image. The Free-Energy-Based Source Identification (FSI)\cite{fsi} method predicts quality by first predicting the reference and degraded images using a sparse representation to model the brain's internal generative model, and then computing the inconsistency between entropies.

General-purpose NR-IQA methods aims to extract quality-aware features from test images for quality estimation \cite{Chen_2022}\cite{Wang_2023}\cite{Ou_2022}\cite{Pan_2023}\cite{Zhu_2022}\cite{Liu_2023}. In DIIVINE\cite{diivine}, to anticipate picture quality degradations, the framework first determined the kind of distortion and then used a regression approach tailored to that distortion. A regression module is then used to map these features to the final score. BRISQUE~\cite{brisque} extracts features based on statistics of natural scene patches in the spatial domain, resulting in 36 features. BLIINDS~\cite{bliinds} employs discrete cosine transform (DCT) to extract contrast and structural features from the image. 

The conventional NR IQA methods have shown promising performance in assessing image quality under various conventional distortions such as Blur, noise, and JPEG-blocking artifacts. Indeed, most general-purpose IQA methods perform fairly well on distortions that impact image structures but degrade significantly when it comes to contrast distortion. Therefore, researchers
have developed specialized methods for evaluating the quality of images that are affected by contrast distortion.

Contrast, which refers to the ability to differentiate objects in an image, is a crucial determinant of image quality and it is considered in a variety of vision tasks \cite{Shafaghi_2023}, and IQA in the presence of contrast change is important in many imaging tasks. For example, a contrast-specific IQA method can be applied to medical image modalities to tune hyper-parameters of contrast enhancement algorithms for diagnosis tasks~\cite{Malekmohammadi_2023}\cite{Malekmohammadi_2024}. Unfortunately, factors such as improper imaging devices and unfavorable environmental conditions can lead to changes in image contrast and consequently result in a decrease in quality.
Some contrast-specific methods are introduced in the FR and RR domains. ~Wang et al.~\cite{pcqi} proposed an FR-IQA patch-based Contrast Quality Index (PCQI) method, which represents image patches in terms of structural components, strength, and average brightness. It calculates the average differences of these components between the distorted and the reference images. Gu et al.\cite{riqmc} is an RR-IQA provided two databases specifically for contrast distortion and proposed a Semi-Reference Image Quality Metric (RIQMC) based on entropy features from the reference image for quality assessment. As mentioned earlier, most of the existing NR-IQA methods are designed for distortions such as blur and JPEG compression, and there has been limited research on evaluating contrast distortion. Among the proposed contrast-specific NR-IQA methods, several methods such as NIQMC\cite{NIQMC}, BCIQM\cite{bciqm}, Fang et al. \cite{fang}, and Zhou et al.~\cite{Zhou} are noteworthy. Based on modeling through NSS, Fang et al.~ \cite{fang} extract
features like Mean, standard deviation, Kurtosis, Skewness, and entropy from images and model
them using Gaussian distributions. Then, features are trained using a support vector regression (SVR). Zhou et al.~\cite{Zhou} transform the image into the LAB color space and extract features in three
domains: spatial luminance, spatial distribution, and orientation, which are trained using a Back Propagation Neural Network (BPNN). Gu et al. \cite{NIQMC} proposed a metric for contrast distortion based on both global and local features. Local contrast distortion is assessed by computing the entropy differences of regions with saliency information, while global contrast distortion is determined by calculating the symmetric Kullback-Leibler divergence between the histograms of the distorted image and uniform distribution. Liu et al. presented a Blind Contrast-distorted Image Quality Model (BCIQM)\cite{bciqm} that extracts features from images in four domains: spatial domain, histogram domain, Chrominance, and Visual Perception. These features are then used to build a quality model based on SVR.

In practice, FR-IQA methods perform better because they use a high-quality
reference image for comparison, but reference images are not always available.
Therefore, NR-IQA methods are highly desirable for practical applications. In the
context of contrast distortion, previous researches has mainly focused on FR-IQA
methods, and limited work has been done on assessing contrast distortion. This
article presents a novel metric based on Pseudo Reference-Image and Classification of Contrast Enhancement methods (PRICCE) to address this issue, where the NR-IQA problem
is transformed into an FR-IQA problem by generating pseudo-reference images
using contrast enhancement algorithms. The idea is to use the most optimum image
enhancement approach to generate an enhanced image from the distorted one that
can be used as a reference for FR IQA. Therefore, the NR problem is transformed
into a FR evaluation. To this end, a classifier network is trained to select the best
contrast enhancement algorithm for a given contrast-distorted image, aiming to
produce an enhanced image that closely matches the reference image. The classifier
learns to choose the most optimum enhancement technique considering the
distortion type, distortion level, and scene statistics. This enhanced image is
then evaluated using FR-IQA algorithms. Extensive experiments on standard
databases with contrast-distorted images, including CCID2014, TID2013, and
CSIQ, demonstrates that the proposed method performs well in assessing this type
of distortion.\\
The main contributions of the proposed method are summarized as follows:

\begin{itemize}
    \item The NR-IQA problem for contrast-distorted images is transformed into an
FR-IQA problem by generating pseudo-reference images using contrast
enhancement algorithms.

    \item A classifier network is deployed to produce pseudo-reference images.

    \item A large dataset is generated for training the classifier network.
\end{itemize}

The article continues with a detailed discussion of the proposed method in Section
~\ref{sec2} and presents experimental results and necessary analyses in Section ~\ref{sec3}, followed
by conclusions in Section ~\ref{sec.conclusion}.

\section{Proposed Method}\label{sec2}
The main idea of this paper is to improve the quality of contrast-distorted images
using contrast enhancement algorithms and then evaluate the quality with
enhanced images. The assumption is that the enhanced images have a quality close
to reference images. However, there is no comprehensive algorithm that can
effectively handle all contrast distortions, such as overexposure and
underexposure, as their performance heavily depends on the image content,
lighting conditions, and the type of contrast distortion applied. Therefore in this paper,
multiple contrast enhancement algorithms are employed to generate pseudo-reference images that closely match the reference image quality. Subsequently,
the pseudo-reference images are compared to the distorted image using full-reference image quality assessment methods and the best algorithm is selected. To find the optimum algorithm during the test when a reference image is not available, a classification network is trained to choose the quality improvement algorithm for a specific
contrast-distorted input image. The network is trained using a dataset consisting of
49,500 contrast-distorted images, each labeled as one of the enhancement algorithm that produces the highest-quality output among the selected algorithms.

\begin{figure}[!]
\centering
\includegraphics[scale=0.35]{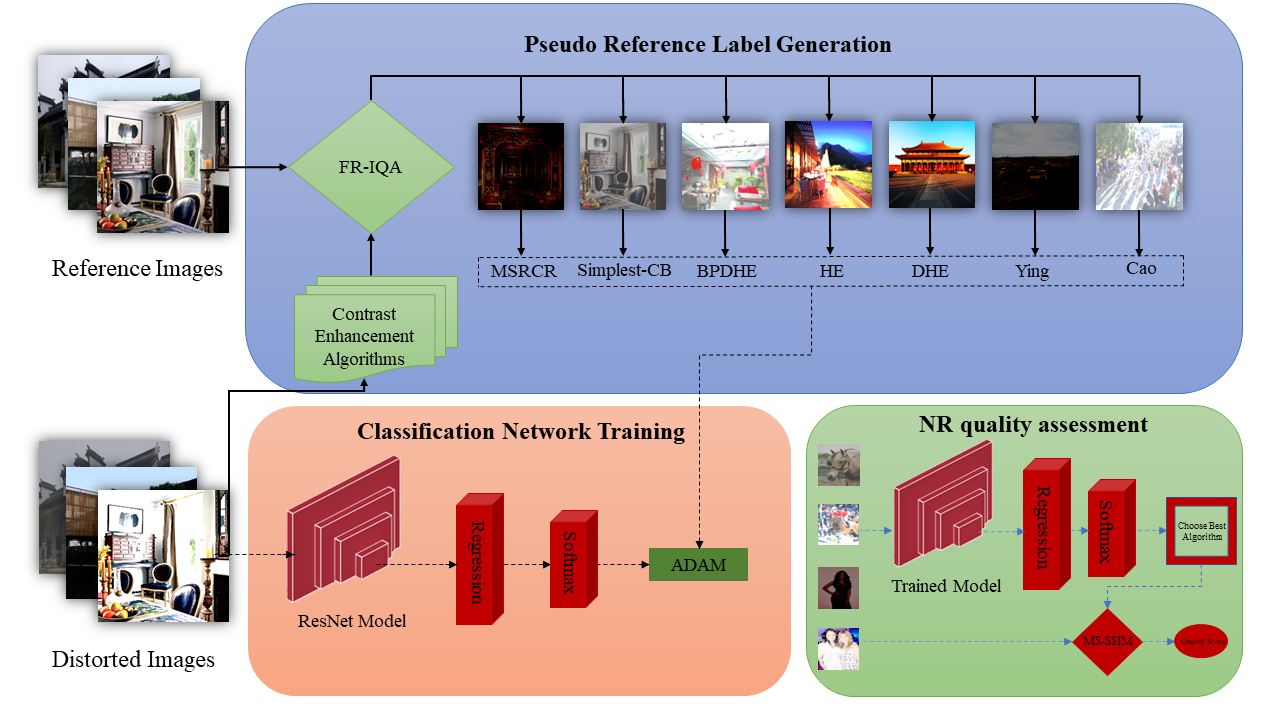}
\caption{Overview of proposed PRICCE}\label{fig1}
\end{figure}

Figure \ref{fig1} shows the overview of the proposed method. The training
dataset is crucial, as it enables the network to learn which enhancement algorithm is likely to provide the best quality output for a given contrast-distorted image. The
paper elaborates further on the building dataset and the methodology for selecting these enhancement algorithms in the subsequent sections.

\subsection{Contrast Enhancement Algorithms}\label{subsec1}
We selected seven widely-used contrast enhancement algorithms based on their performance on different types of contrast distortion. The utilized algorithms include Histogram
Equalization (HE), Dynamic Histogram Equalization (DHE)\cite{dhe}, Brightness
Preserving Dynamic Histogram Equalization (BPDHE)\cite{bpdhe}, Simplest Color Balance\cite{simplest-cb},
Multiscale Retinex Algorithm with Color Restoration (MSRCR)\cite{msrcr}, Ying et al.~\cite{Ying_2017}, and Cao et al.~\cite{Cao_2018}.\\
HE is a common method used to enhance the contrast of an image by spreading and equalizing the image histogram across the entire brightness range.

\begin{equation}\label{eq1}
   s_ {k}  =T( r_ {k} )=(L-1)  \sum _ {r=0}^ {k}  p_ {r}  (  r_ {j}  ) \quad where \quad k=0,1,  \cdots  ,L-1 
\end{equation} Where $s$ and $r$ are the output and input pixel intensity, respectively. $k$ is the intensity level ranging from $0$ to $L$  and $T(.)$ is the transformation function. $p_r(r_j)$ is the probability of occurrence of intensity level $r_j $ in an image.  \\
In DHE, the histogram is first divided into multiple partitions based on local
minima. Then, each partition is assigned a gray level range. Subsequently, the HE
method is applied to each section.
\begin{equation}\label{eq2}
    \begin{split}
   span_ {i} = m_ {i} - m_ {i-1} \\
   range_ {i} = \frac {span_ {i}}{\sum{span_ {i}}} *(L-1)
   \end{split}
\end{equation}
 
Where $span_i$, $m_i$, and $range_i$ are the dynamic gray level range of the input image, local minima, and
dynamic gray level range of the output image at $i$th sub-histogram.\\
The BPDHE method is an extension of the DHE, aiming to generate an output image with an intensity mean equal to the input image's mean. It smooths the histogram, partitions it, maps each partition to a new dynamic range and normalizes the output intensity. This method improves the enhancement and preservation of average brightness compared to DHE. The method interpolates missing brightness levels, smooths the histogram, detects local maxima, maps partitions to new dynamic ranges, and finally normalizes the image as follows:
\begin{equation}\label{eq3}
    g(x,y)=( \frac {M_ {i}}{M_ {o}})f(x,y)
\end{equation}

Where $g(x,y)$ is the final output and $f(x,y)$ is the output just after the equalization process. $M_ {i}$ and $M_ {o}$ are mean brightness of input and output respectively.\\

Contrast distortion caused by underexposure, or artificial or natural lighting conditions such as sunset could be corrected by color balance algorithms like Simplest Color Balance. These techniques depend on the idea that white must be represented by the greatest values of R, G, and B presented in the image, and obscurity by the lowest values. The three color channels are simply stretched as far as possible by the algorithm to fill the whole range [0, 255]. The simplest way to accomplish this is to apply an affine transform $x: Ax+B$ to each channel, then calculate $A$ and $B$ to make the maximal value in the channel 255 and the minimum value 0 \cite{simplest-cb}.\

The MSRCR method, based on the Retinex theory, is introduced to separate the reflectance component from images. This method includes a color restoration factor for each color channel.
\begin{equation}\label{eq4}
    MSRCR=  \log R_ {i}  (x,y)=  \sum _ {k=1}^ {N} C_ {i}  \omega _ {k} {\log I_i(x,y)- \log [G_{k}(x,y)* I_ {i}(x,y)]}
\end{equation}
Where $C$ and $log(.)$ represents the restoration factor of the $i$th color channel and logarithm operation, $G_k$ is the Gaussian function at $k$ Gaussian scale, $N$ is the number of scales and $\omega$ represents the weight of each scale. The parameter $I(x,y)$ represents the input image intensity at each pixel location $(x,y)$.\\
Ying et al.~\cite{Ying_2017} proposes an image contrast enhancement algorithm using illumination estimation techniques. It involves formulating a weight matrix for image fusion, introducing a camera response model, synthesizing multi-exposure images,
determining the optimal exposure ratio, and fusing the input and synthetic images
based on the weight matrix to achieve precise contrast enhancement. The fused
image $R^C$ is defined as:
\begin{equation}\label{eq5}
 R^ {C} = W\circ P^ {C} + (1-W)\circ g(P^ {C},k)
\end{equation}
Where $W$ is estimated weight matrix, $g$ is called Brightness Transform Function (BTF), $k$ is
exposure ratio and $P^ {C}$ is input image $p$ at color channel $C$.\\
Cao et al.~\cite{Cao_2018} have employed adaptive gamma correction (AGC) to enhance images, utilizing
negative images for bright images and Truncated cumulative distribution function for dimmed images as input to AGC. The transformed pixel intensity using AGC is:
\begin{equation}\label{eq6}
T(l)= round[l_{\max}(\frac {l}{l_ {\max }})^ {\gamma(l)}]
\end{equation}
Where $\gamma(l)$ is the CDF of gray levels in the input image, $l$ is the intensity level, and $round(.)$ is the rounding operation. Finally, a weighting distribution function is used to smooth the primary histogram.\\

The need to choose between various enhancement algorithms to obtain the highest quality reference arises from the limitations of these algorithms under different conditions. No single algorithm can perform well on different contrast distortions and image characteristics. Algorithms that operate in the histogram domain, such as DHE, BPDHE, and HE, for example, can provide decent enhanced images for different contents but can also suffer from issues like over-enhancement. Therefore, other algorithms were also used to mitigate these problems. Figure \ref{fig2} displays examples of a reference image, a degraded image, and contrast-enhanced images from the CCID2014 database.

\subsection{Dataset Generation}\label{subsec2}
The database intended for training the classification network should include
a large number of degraded images each with a class label corresponds to the algorithm that
have the best quality output among other algorithms. To generate these
degraded images, the Waterloo database\cite{waterloo}, consisting of 4744 high-quality
images were utilized. In the initial step, 1500 images were selected from
this database to apply synthetic contrast distortions. The applied degradation types include contrast change, gamma transfer, cubic and logistic function, and mean-shifting, as outlined in Table \ref{tab1}.

\begin{table}[h] 
\caption{Types of contrast distortions used to generate dataset}\label{tab1}%
\begin{tabular}{@{}c|c@{}}
\toprule
\textbf{Contrast Distortion Type} & \textbf{No. of Distortion Levels}\\
\midrule
Contrast Change   & 5  \\
Gamma Transfer    & 8   \\
Logistic Function & 4    \\
Cubic Function & 4    \\
Mean-Shifting & 12    \\
\botrule
\end{tabular}
\end{table}

\begin{itemize}

    \item\textbf{\textit{Contrast Change}:} Contrast reduction is obtained by blending the input image with a gray image. The equation can be defined as:
    
    \begin{equation}\label{eq7}
    I_{\text {dist}}^{{cc}}=(1-\alpha) * I_{{ref }}{ }^{rgb}+\alpha * I_{{ref }}{ }^{gray} \quad \text { \quad where \quad} \alpha=\{0.5,0.75,1,1.25,1.5\}
\end{equation}

    \item \textbf{\textit{Gamma Transfer}:} is a non-linear adjustment to individual pixel values. It is
simply a power law function which is defined as
    \begin{equation}
    I_{\text {dist}}^{{\gamma}}=I_{\text {ref }}{ }^{\left(\frac{1}{\gamma}-1\right)} \text { where } \gamma=\left\{\frac{1}{5}, \frac{1}{3}, \frac{1}{2}, \frac{1}{1.5}, 1.5,2,3,5\right\}
    \end{equation}

    \item \textbf{\textit{Cubic and Logistic Functions}:} applies 3-order cubic and 4-parameter logistic functions on reference images. The equations are:
    \begin{equation}\label{cubiceq}
    I_{\text {dist}}^{{cubic}}=a_1 I_{\text {ref }}^3+a_2 I_{\text {ref }}^2+a_3 I_{\text {ref }}+a_4 
    \end{equation}
        
    \begin{equation}\label{logisticeq}
    I_{\text {dist}}^{{logistic}}=\frac{b_1-b_2}{1+e^{\left(-\frac{x-b_3}{b_4}\right)}+b_2}
    \end{equation}
where $\alpha$ and $\beta$ are obtained by leveraging the method used in \cite{Gu_2016_dist}.

    \item \textbf{\textit{Mean-Shifting}:} all intensity values of the reference image is shifted by 12 different levels, defined as:
        \begin{equation}\label{eq11}
        I_{\text {dist }}^{\text {Mean-shift }}=I_{\text {ref }}+\Delta \quad \text {\quad where \quad} \Delta=\{ \pm 20, \pm 40, \pm 60, \pm 80, \pm 100, \pm 120\}
        \end{equation}

\end{itemize}

\begin{figure}[!]
\centering
\includegraphics[scale=0.45]{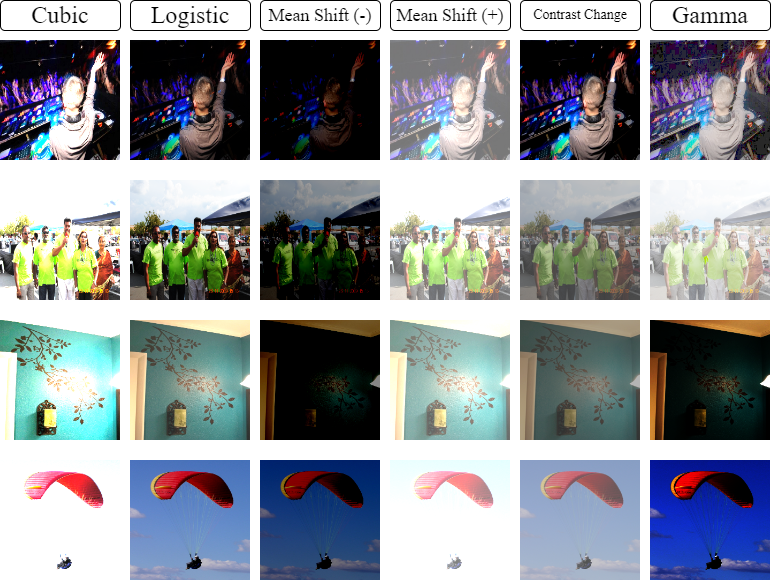}
\caption{Samples of the generated dataset in different types and levels of distortions}\label{sample}
\end{figure}

\begin{figure}[!]
\centering
\includegraphics[scale=0.45]{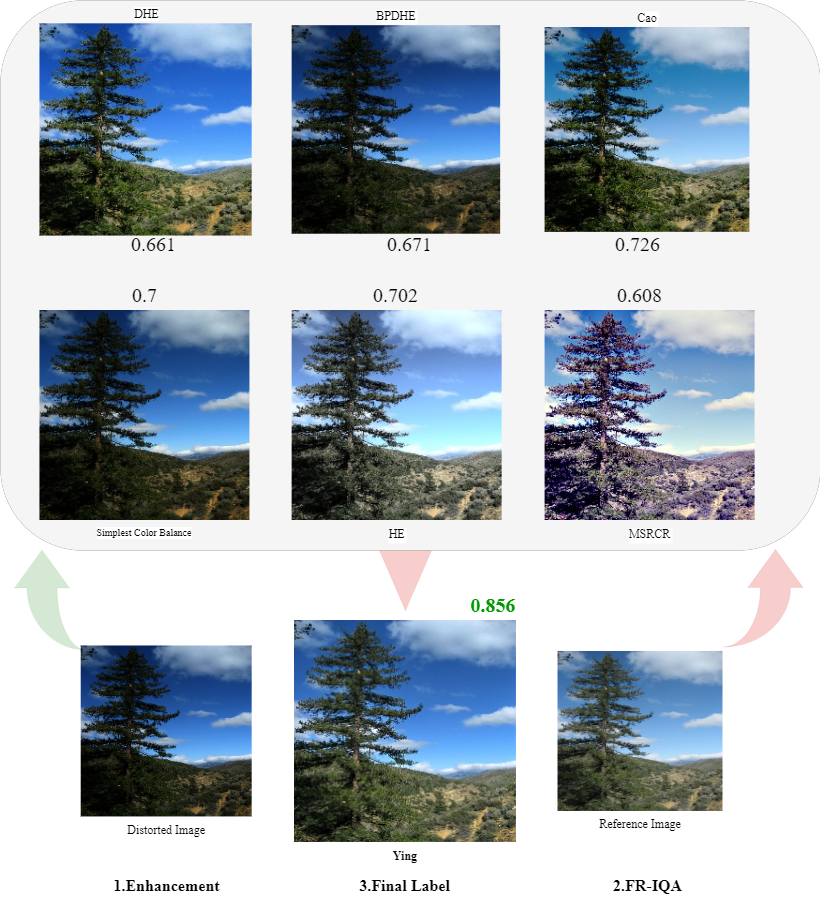}
\caption{Labelling process of the generated dataset}\label{fig2}
\end{figure}

As shown in Table~\ref{tab1}, the total number of applied degradations levels is 33 applied to 1500 images, meaning that the database consists of a total number of 49,500 images. Figure \ref{sample} presents examples of the generated distorted images at different types and distortion levels. The next step of the preparation is to label these images. For this purpose, each degraded image is processed by the 7 selected enhancement algorithms introduced in the previous section. The output of each contrast improvement algorithm is compared to the reference image using FR metrics. The effectiveness of VIF has been proved in several studies for different distortions including contrast \cite{Mason_2020}\cite{Hassan_2022}\cite{Hassan_2022_1}\cite{Nair_2021}, thus it was selected in our work. Finally, the enhancement algorithm whose output image showed the highest visual quality, or in other words, was closest in quality to the reference image, was selected as the label of the degraded image. Figure \ref{fig2} illustrates the labeling process.

\subsection{Classification Network}\label{subsec3}
Based on the input distorted images and the associated labels, a classification network was trained to automatically select the optimum contrast enhancement algorithm, that can produce the highest-quality image from the input. 

We selected ResNet18 which offers good accuracy and speed as the classification network.  The network include 18 layers and pre-trained on the ImageNet database. Modifications were made to the final layers of this network, including adding two fully connected layers with
256 and 128 neurons at the end of the network and a Dropout layer with a 0.5 probability to prevent overfitting. Additionally, the number of classes in this
network is set to 7, corresponding to the number of contrast enhancement algorithms. 

The trained network is used to select the optimum algorithm for image enhancement. The next step is to produce a pseudo-reference image using the network-selected enhancement algorithm. The final step is to evaluate the enhanced pseudo-reference image against the degraded input using FR-IQA.

%\begin{table}[!hbp]
\begin{sidewaystable}
\caption{Comparison of proposed PRICCE method with different methods.}\label{tab4}
\hspace*{-15cm}
\centering
\renewcommand{\arraystretch}{1.5}
\begin{tabular}{|c|c|cccc|cccc|cccc|}
\hline
\multirow{2}{*}{\begin{turn}{90}IQA Type\end{turn}} & \multirow{2}{*}{Method} & \multicolumn{4}{c|}{CSIQ} & \multicolumn{4}{c|}{TID2013} & \multicolumn{4}{c|}{CCID2014} \\
 & & \begin{turn}{90}SROCC\end{turn} & \begin{turn}{90}KROCC\end{turn} & \begin{turn}{90}PLCC\end{turn} & \begin{turn}{90}RMSE\end{turn} & \begin{turn}{90}SROCC\end{turn} & \begin{turn}{90}KROCC\end{turn} & \begin{turn}{90}PLCC\end{turn} & \begin{turn}{90}RMSE\end{turn} & \begin{turn}{90}SROCC\end{turn} & \begin{turn}{90}KROCC\end{turn} & \begin{turn}{90}PLCC\end{turn} & \begin{turn}{90}RMSE\end{turn} \\
\hline
\multirow{6}{*}{\begin{turn}{90}FR-IQA\end{turn}}  & PSNR\cite{psnr} & 0.908 & 0.675 & 0.925 & 0.064 & 0.445 & 0.310 & 0.442 & 1.025 & 0.687 & 0.494 & 0.607 & 0.514\\
 & SSIM\cite{ssim} & 0.779 & 0.589 & 0.801 & 0.093 & 0.319 & 0.235 & 0.543 & 0.942 & 0.811& 0.607 & 0.818 & 0.373\\
 & PCQI\cite{pcqi} & 0.940 & 0.818 & 0.955 & 0.048 & 0.876 & \textbf{0.730} & \textbf{0.947} & \textbf{0.158} & 0.861 & 0.674 & 0.888 & 0.290\\
 & FSIM\cite{fsim} &0.942 & 0.788& 0.950& N/A& 0.439& 0.360& 0.657& N/A& 0.765& 0.570& 0.820 &N/A\\
 & VSI\cite{vsi} &\textbf{0.950} &0.809 &0.953 &0.050 & 0.464& 0.370&0.678 &0.767 &0.733 &0.573 &0.791 &0.373\\
 & VIF\cite{vif} & 0.934 & N/A & 0.943& N/A & 0.771& N/A& 0.845& N/A& 0.834& N/A& 0.858&N/A\\
\hline
\multirow{3}{*}{\begin{turn}{90}RR-IQA\end{turn}}  & RRED\cite{rred} & 0.926& 0.790& 0.945& 0.052& 0.390& 0.315& 0.615& 0.886& 0.658& 0.471& 0.708&0.458\\
 & RIQMC\cite{riqmc} & 0.949& 0.836& 0.890& 0.068& 0.841& 0.649& 0.868& 0.322& 0.847& 0.661& 0.843&0.621\\
 & FSI\cite{fsi} & 0.945& 0.828& \textbf{0.964}& \textbf{0.042}& 0.518& 0.384&0.770&0.757 & 0.598&0.426 &0.727 &0.446\\\hline
\multirow{7}{*}{\begin{turn}{90}\shortstack{General Purpose \\ NR-IQA}\end{turn}}  & BIQME\cite{biqme}   & 0.768&0.596 &0.776 &0.091 & 0.783& 0.594& 0.851& 0.599& 0.827& 0.576& 0.800&0.389\\
 & BIQI\cite{biqi} & 0.83& \textbf{0.849}&0.650 & 0.087& 0.325& 0.226& 0.375& 0.897& 0.584& 0.416&0.616 &0.517\\
 & HOSA\cite{hosa} &0.693 & N/A& 0.778& N/A& 0.409& N/A& 0.410&N/A & 0.618&N/A &0.711 &N/A\\
 & GMLOG\cite{gmlog} & 0.767& N/A & 0.797& N/A & 0.433&  N/A&0.543 & N/A &0.739 & N/A &0.807 & N/A\\ 
 & BLIINDS\cite{bliinds} &0.490 & 0.358&0.551 &0.137 & 0.223& 0.155& 0.267& 0.927& 0.446& 0.311& 0.492&0.567\\
 & BRISQUE\cite{brisque} &0.454 &0.320 &0.496 &0.564&0.323 &0.226 &0.388 &0.886 &0.641 &0.478 &0.665 &0.423\\
 & DIIVINE\cite{diivine} &0.708 &0.519 &0.744 &0.519 &0.287 &0.199 & 0.330& 0.917&0.373 &0.258 &0.438 &0.580\\ \hline
\multirow{6}{*}{\begin{turn}{90}\shortstack{Contrast-specific\\ NR-IQA}\end{turn}}  & NIQMC\cite{NIQMC} &0.853 &0.669 &0.864 &0.085 &0.646 &0.469 &0.723 &0.678 &0.811 &0.605 &0.844 &0.351\\
 & Fang\cite{fang} &0.801 &0.608 &0.822 &0.094 &0.502 &0.419 & 0.664&0.826 &0.805 &0.598 &0.840 &0.369\\
 & BCIQM\cite{bciqm} &0.840 &0.660 &0.865 &0.081 &\textbf{\textcolor{red}{0.881}} &\textcolor{red}{0.728} &\textcolor{red}{0.926} &0.419 &0.885 &0.706 &0.896 &0.287\\
 & Zhou\cite{Zhou} &0.888 &0.729 &0.937 &0.052 &0.840 &0.661 &0.896 &\textcolor{red}{0.386} &\textbf{\textcolor{red}{0.912}} &\textbf{\textcolor{red}{0.724}} &\textbf{\textcolor{red}{0.924}} &\textbf{\textcolor{red}{0.267}}\\
 & MDM\cite{mdm} &0.928 & N/A&0.904 & N/A&0.851 & N/A&0.912 & N/A&0.836 & N/A&0.871 &N/A\\
 & \textbf{\shortstack{PRICCE(prop.)}} &\textcolor{red}{0.940} & \textcolor{red}{0.789}& \textcolor{red}{0.953}& \textcolor{red}{0.050}& 0.759&0.585 &0.768 &0.627 &0.811 &0.606 &0.825 &0.369\\
\hline
\end{tabular}
\hspace*{-15cm}
%\end{table}
\end{sidewaystable}

\section{Experimental Results}\label{sec3}

The performance evaluation of the proposed metric is presented in this section. Three subjectively-annotated databases, CCID2014, CSIQ, and TID2013, were used in the evaluation, containing contrast-distorted images. The CCID2014 database consists of 655 contrast-degraded images, with several contrast change methods,
applied to 15 reference images. The types of contrast alterations in CCID2014
include Gamma Transfer, Convex and Concave arcs, Mean-shifting, Cubic and
Logistic Functions, and Compound Functions. The TID2013 database includes 3000 distorted images and 25 reference images which 24 types of distortion in 5 levels are applied on each reference image. We used images with contrast change and mean shift distortions consisting of 250 images in our evaluation. The CSIQ database
contains 116 contrast-distorted images with global contrast decrements  at five different levels, that the associated DMOS (Differential Mean
Opinion Score) for quality assessment. \\
To evaluate the performance of the proposed method, a comparison was made with general-purpose FR-IQA (PSNR\cite{psnr}, SSIM\cite{ssim}, FSIM\cite{fsim}, VSI\cite{vsi} and VIF\cite{vif}), RR-IQA (RRED\cite{rred},  and FSI\cite{fsi}), and NR-IQA (BIQME\cite{biqme}, BIQI\cite{biqi}, HOSA\cite{hosa}, GMLOG\cite{gmlog}, BLIINDS\cite{bliinds}, BRISQUE\cite{brisque} and DIIVINE\cite{diivine}) methods, as well as methods specially designed for contrast-distorted images, including PCQI\cite{pcqi}, RIQMC\cite{riqmc}, Fang\cite{fang}, Zhou\cite{Zhou}, NIQMC\cite{NIQMC}, BCIQM\cite{bciqm} and MDM\cite{mdm}.
To compare the performance, four common evaluation criteria were used. Two criteria, Kendall’s Rank
Order Correlation Coefficient (KROCC) and Spearman’s Rank Order Correlation
Coefficient (SROCC), assesses prediction monotonicity. Two other metrics, Pearson
Linear Correlation Coefficient (PLCC) and Root Mean Square Error (RMSE), evaluate prediction accuracy. It is recommended by the Video Quality Expert Group (VQEG) to apply a non-linear regression function to the estimated scores of an
objective IQA metric before calculating PLCC and RMSE to remove nonlinearity.
Therefore, the \textit{nlinfit} function in MATLAB was used for this purpose. For an
Objective IQA metric, higher values of SROCC, PLCC, and KROCC close to 1 are
expected, while RMSE should tend towards zero. 

In the first step, we report the outcomes of the deep network used to classify images into one of the seven image enhancement algorithms. The number of images assigned to each class among all training images is shown in Table \ref{tab2}. As can observed, the classes ‘HE’ and ‘Simplest Color Balance’ have more images assigned to them than other classes, indicating their effectiveness but there are still a significant number of images that are better enhanced using the other five algorithms.

\begin{table}[h] 
\caption{Number of images per class in the generated dataset}\label{tab2}%
\begin{tabular}{@{}cc@{}}
\toprule
\textbf{Contrast Enhancement} & \textbf{No.of Labels}\\
\midrule
HE& 9378    \\
Simplest Color Balance\cite{simplest-cb}   & 10864  \\
Ying\cite{Ying_2017}    & 3988   \\
Cao\cite{Cao_2018} & 3720    \\
DHE\cite{dhe} & 2303    \\
BPDHE\cite{bpdhe} & 1246    \\
MSRCR\cite{msrcr} & 2660    \\
\botrule
\end{tabular}
\end{table}

\begin{figure}[!]
\centering
\includegraphics[scale=0.55]{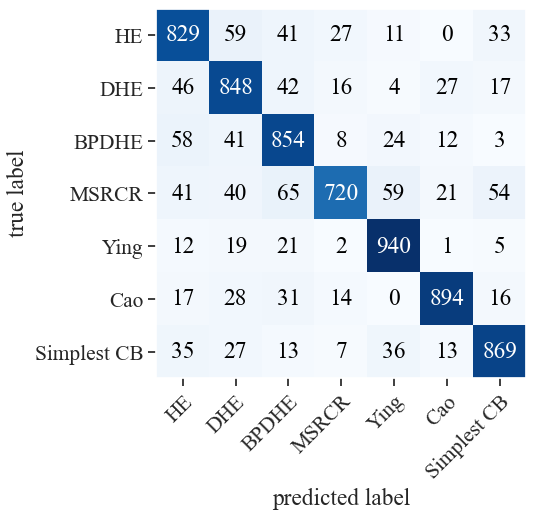}
\caption{Confusion matrix of image enhancement algorithms}\label{confusionmatrix}
\end{figure}

The classification network was implemented using the PyTorch framework. Since the generated database is imbalanced, class weighting was used, which optimizes the importance of classes with fewer data points. The input image size to the network was set to 224$\times$224. The optimizer used is of the ADAM type, with an initial learning rate is set to 0.001, which is reduced by a factor of 0.1 at every 50 epochs. The batch size was set to 128 and ReLU activation functions were used. Data augmentation techniques were also applied to prevent overfitting. It is important to note that the augmentation techniques should not alter the color and structure of the training images since they could affect the assigned labels. For example, using a color jitter on training images cannot be done as it changes the color, making it incompatible with the assigned labels. Therefore, data augmentation techniques that do not affect the color or structure of the images, such as 45-degree rotation, and horizontal and vertical flip, were applied.\\
We used 7000 images for the test (1000 images per class) in which 85\% classification accuracy was achieved over 150 epochs with stopping criteria.) The confusion matrix of the classification results has been shown in Figure \ref{confusionmatrix}. As shown, the classification network can do a fairly good job of identifying the correct labels considering that some enhancement methods have rather close visual outcomes.

\begin{table}[!hbp]
\caption{Performance of the proposed method integrated with different FR-IQA approaches.}\label{tab3}
\begin{tabular}{c|c|c|c|c|c|c}
\toprule
Dataset & Criteria & \multrow{Proposed+\\SSIM} & \multrow{Proposed+\\MS-SSIM} & \multrow{Proposed+\\VIF} & \multrow{Proposed+\\DIST} & \multrow{Proposed+\\GMSD}\\
\midrule
\multirow{4}{*}{CSIQ} & SROCC & 0.908 & \textbf{0.940} & 0.926 & 0.938 & 0.711 \\
    & KROCC & 0.736 & \textbf{0.789} & 0.758 & 0.781 & 0.534 \\
    & PLCC & 0.900 & \textbf{0.953} & 0.930 & 0.939 & 0.744 \\
    & RMSE & 0.109 & \textbf{0.050} & 0.061 & 0.057 & 0.112 \\ \hline
 \multirow{4}{*}{TID2013} & SROCC & 0.663 & \textbf{0.759} & 0.743 & 0.631 & 0.451 \\
    & KROCC & 0.499 & \textbf{0.585} & 0.559 & 0.484 & 0.326 \\
    & PLCC & 0.640 & 0.768 & \textbf{0.808} & 0.757 & 0.364 \\
    & RMSE & 0.74 & 0.627 & \textbf{0.577} & 0.640 & 0.913 \\ \hline   
 \multirow{4}{*}{CCID2014} & SROCC & 0.782 & \textbf{0.811} & 0.747 & 0.710 & 0.644 \\
    & KROCC & 0.576 & \textbf{0.606} & 0.549 & 0.518 & 0.465 \\
    & PLCC & 0.796 & \textbf{0.825} & 0.772 & 0.729 & 0.680 \\
    & RMSE & 0.395 & \textbf{0.369} & 0.415 & 0.447 & 0.478 \\ 
\botrule
\end{tabular}
\end{table}

\begin{figure}[!]
\centering
\includegraphics[scale=0.35]{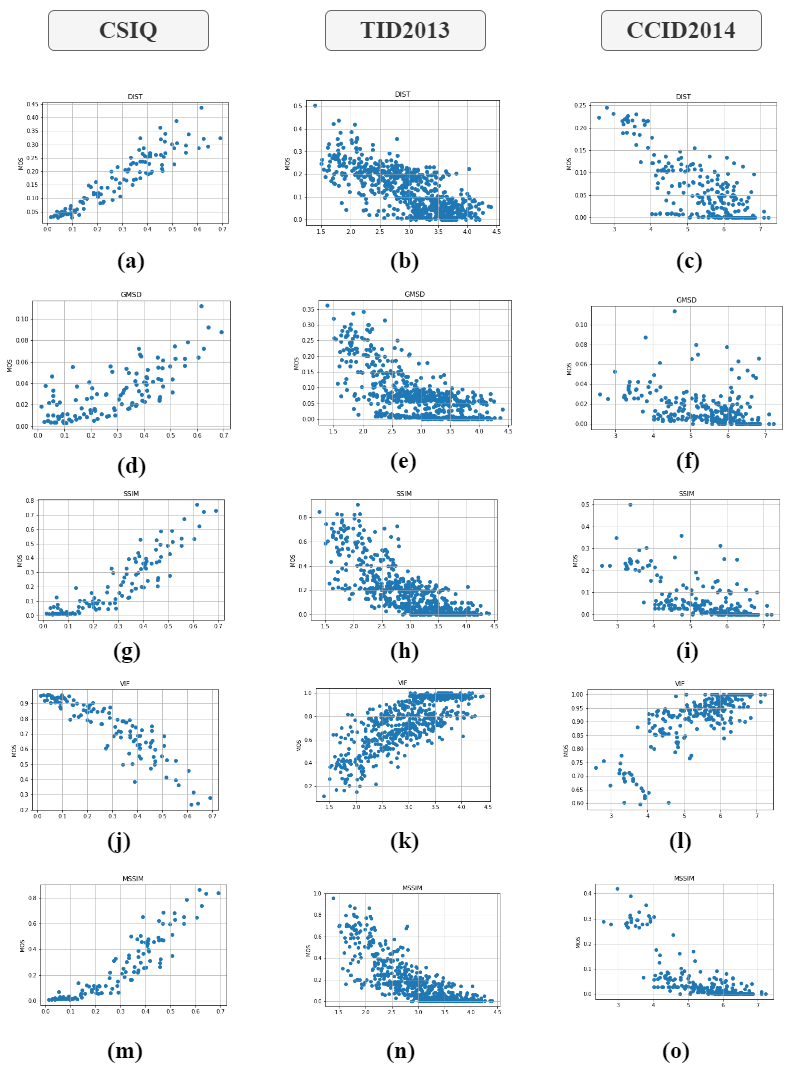}
\caption{Scatter plots of MOS vs proposed PRICCE scores. Pseudo-references and distorted images are compared under different FR-IQA methods. (a-c) DIST (d-f) GMSD (g-i) SSIM (j-l) VIF (m-o) MS-SSIM}\label{scatterplots}
\end{figure}

After the classification network found the most suitable enhancement algorithm, the selected method was used to make a pseudo-reference from the input contrast-distorted image. Next, the quality change between the pseudo-reference and the distorted image can be measured using any FR metric. Here we examined the performance of five widely-used FR-IQA methods including GMSD, DIST, SSIM, VIF, and MS-SSIM. Table \ref{tab3} presents the performance evaluation results of the proposed method when integrated with different FR metrics. Due to the overall superior performance of the MS-SSIM method compared to the others, it was integrated with the proposed method.

Table \ref{tab4} compared the performance of the proposed method with all competing metrics on three image datasets. The best metric values among FR, RR, general-purpose NR, and contrast-specific NR IQA metrics are shown in bolded text as well as best performance in contrast-specific NR-IQA is highlighted in red color. As it can be observed, our method, PRICCE, almost performs better than all general-purpose NR-IQA methods and can outperform or perform comparably to several FR-IQA methods. In comparison to the RR-IQA methods, our metric only had slightly lower performance compared to the RIQMC method, given that our evaluation is conducted without the use of reference images. Finally, in comparison with Contrast-specific NR-IQA
methods, PRICCE outperformed the NIQMC and Fang methods, achieving better performance
than other methods, especially on the CSIQ database. Scatter plot diagrams of the scores obtained from the proposed method against MOS scores are shown in Figure \ref{scatterplots}, indicating a good monotonic relationship between the proposed
method's scores and the subjective scores. 

%\textcolor{red}{We performed a statistical significance test using the SROCC index of six best-performing metrics on the CSIQ datasets, according to ITU-T Recommendation P.1401 For each pair of metrics, we conducted a two-sample student t-test with 95\% confidence level. The results are presented in a table, where a symbol ‘1’ indicates that a metric in the row axis is statistically superior to the method on the column, ‘-1’ indicates that the row metric is inferior to the column metric, and the value ‘0’ means there is no statistical significance in terms of SROCC. Our proposed PRICCE method was found to be statistically superior to other competing methods}

\section{Conclusion}\label{sec.conclusion}
This paper proposes a No-Reference Image Quality Assessment (NR-IQA)
method specialized for contrast distortion. To evaluate image quality, several
contrast enhancement algorithms were used to generate pseudo-reference images.
However, since a single algorithm cannot provide the best results for all contrast
distortions, a classification network was utilized to select an algorithm that closely
matches the quality of the reference image. To train this network, a database
consisting of 49,500 images was created, with labels representing the algorithm
that performed best compared to other selected algorithms. Finally, the proposed
method was compared with state-of-the-art methods on three databases: CCID2014, TID2013, and CSIQ. The performance evaluation results showed the effectiveness of the proposed method under different contrast distortions.

% \section*{Data Availability}
% All evaluated datasets are publicly available. The dataset used for classification can be generated by equations \ref{eq7} to \ref{eq11} applied on the Waterloo dataset. The code to generate this dataset is available \href{https://github.com/hamid-mp/Image-Enhacement-and-Quality-Assessment}{here}.

\section{Declarations}\label{sec.declare}

\subsection{Funding}
Not applicable

\subsection{Conflict of Interest}
Not applicable

\bibliography{sn-bibliography}


\end{document}